\pdfoutput=1

\documentclass[11pt]{article}

\usepackage[final]{coling}

\usepackage{times}
\usepackage{latexsym}
\usepackage{makecell}

\usepackage[T1]{fontenc}

\usepackage[utf8]{inputenc}

\usepackage{microtype}

\usepackage{inconsolata}

\usepackage{graphicx}
\usepackage{subcaption}
\usepackage{multirow}
\usepackage{tabularx}
\usepackage{booktabs}
\usepackage{amsmath} 
\usepackage{amssymb} 
\usepackage{appendix}
\usepackage[flushmargin]{footmisc} 

\usepackage{times}
\usepackage{latexsym}

\usepackage[T1]{fontenc}

\usepackage[utf8]{inputenc}

\usepackage{microtype}

\usepackage{inconsolata}

\usepackage{acronym}
\usepackage{enumitem} 

\acrodef{MI}{motivational interviewing}
\acrodef{MISC}{motivational interviewing skills code}
\acrodef{MITI}{motivational interviewing treatment integrity}
\acrodef{LLM}{large language model}
\acrodef{NLP}{natural language processing}
\acrodef{NLG}{natural language generation}
\acrodef{BC}{behavioral coding}
\acrodef{RLHF}{reinforcement learning from human feedback}

\acrodef{BiMISC}{BiMISC}
\acrodef{AnnoMI}{AnnoMI}

\acrodef{QS}{question}
\acrodef{OQ}{open question}
\acrodef{CQ}{closed question}
\acrodef{RF}{reflection}
\acrodef{SR}{simple reflection}
\acrodef{CR}{complex reflection}
\acrodef{TI}{therapist input}
\acrodef{ADV}{advice}
\acrodef{AFF}{affirm}
\acrodef{DIR}{direct}
\acrodef{EC}{emphasize control}
\acrodef{FA}{facilitate}
\acrodef{FIL}{filler}
\acrodef{GI}{giving information}
\acrodef{SP}{support}
\acrodef{STR}{structure}
\acrodef{WAR}{warn}
\acrodef{PS}{permission seeking}
\acrodef{OP}{opinion}

\acrodef{NT}{neutral talk}
\acrodef{ASK}{ask}
\acrodef{FN}{follow/Neutral}
\acrodef{CT}{change talk}
\acrodef{CM+}{commitment+}
\acrodef{TS+}{taking step+}
\acrodef{R+}{reason+}
\acrodef{O+}{other+}
\acrodef{ST}{sustain talk}
\acrodef{CM-}{commitment-}
\acrodef{TS-}{taking step-}
\acrodef{R-}{reason-}
\acrodef{O-}{other-}

\acrodef{CM}{commitment}
\acrodef{TS}{taking step}
\acrodef{R}{reason}
\acrodef{O}{other}

\acrodef{MI}{motivational interviewing}
\acrodef{MITI}{motivational interviewing treatment integrity}
\acrodef{MISC}{motivational interviewing skill code}
\acrodef{CoS}{Chain-of-Strategy}
\acrodef{NLG}{natural language generation}
\acrodef{LLM}{large language model}

\newcommand{\done}[1]{\textcolor{black}{#1}} 

\renewcommand\thesection{\arabic{section}}

\title{Rethinking the Alignment of Psychotherapy Dialogue Generation with Motivational Interviewing Strategies}

\author{
     \textbf{Xin Sun\thanks{Corresponding author. Email: \href{mailto:x.sun2@uva.nl}{x.sun2@uva.nl}}\textsuperscript{1,2}},
     \textbf{Xiao Tang\textsuperscript{3}},
     \textbf{Abdallah El Ali\textsuperscript{2,5}},
     \textbf{Zhuying Li\textsuperscript{3}},
     \\
     \textbf{Pengjie Ren\textsuperscript{4}},
     \textbf{Jan de Wit\textsuperscript{6}},
     \textbf{Jiahuan Pei\textsuperscript{7}},
     \textbf{Jos A.Bosch\textsuperscript{1}} \\
     \textsuperscript{1}University of Amsterdam, the Netherlands;\\
     \textsuperscript{2}Centrum Wiskunde \& Informatica (CWI), the Netherlands;\\
     \textsuperscript{3}Southeast University, China;
     \textsuperscript{4}Shandong University, China;\\
     \textsuperscript{5}Utrecht University, the Netherlands;
     \textsuperscript{6}Tilburg University, the Netherlands; \\
     \textsuperscript{7}Vrije Universiteit Amsterdam, the Netherlands\\ \\
}

\begin{document}

\maketitle


\begin{abstract}
    Recent advancements in large language models (LLMs) have shown promise in generating psychotherapeutic dialogues, particularly in the context of motivational interviewing (MI). However, the inherent lack of transparency in LLM outputs presents significant challenges given the sensitive nature of psychotherapy. Applying \textit{MI strategies}, a set of MI skills, to generate more controllable therapeutic-adherent conversations with explainability provides a possible solution. 
    In this work, we explore the alignment of LLMs with MI strategies by first prompting the LLMs to predict the appropriate strategies as reasoning and then utilizing these strategies to guide the subsequent dialogue generation. We seek to investigate whether such alignment leads to more controllable and explainable generations. Multiple experiments including automatic and human evaluations are conducted to validate the effectiveness of MI strategies in aligning psychotherapy dialogue generation. Our findings demonstrate the potential of LLMs in producing strategically aligned dialogues and suggest directions for practical applications in psychotherapeutic settings.
\end{abstract}



\begin{figure*}[!htbp]
    \centering
    \includegraphics[width=0.985\textwidth]{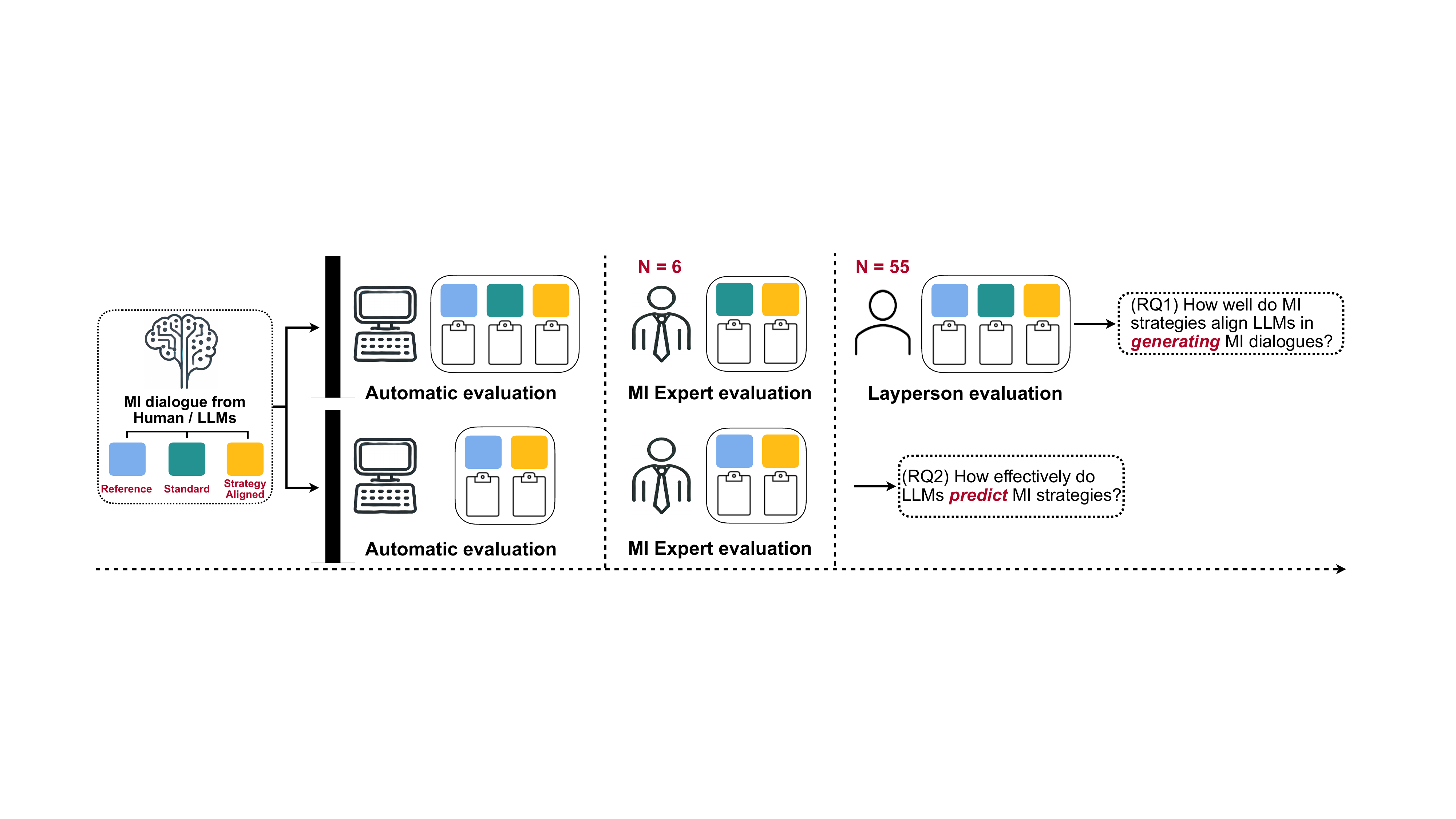} 
    \vspace{-0.0mm}
    \caption{Visual summary of the research questions and corresponding experimental evaluations did in this work.}
    \label{procedure}
    \vspace{-1.0mm}
\end{figure*}

\section{Introduction}

\Acf{MI} is a client-centered counseling technique aimed at encouraging individuals to change behaviors via motivating conversations~\cite{mi-1}. It can boost self-directed (i.e., instead of instructed or coerced) and autonomous motivation for behavior change through collaboration between therapists and clients, by emphatic conversations that address ambivalence about such change, and by enhancing a sense of self-efficacy~\cite{martins2009review}. The effectiveness of MI as a conversational technique is strongly supported by empirical evidence that demonstrates clients' adherence to interventions~\cite{ALPERSTEIN2016393}.
Without the use of MI, traditional techniques can potentially cause resistance and disengagement from clients due to their confrontational ways of thinking~\cite{miller428motivational}.
Key to ensuring MI's effectiveness are strategic schemes, such as 
\acf{MISC}~\cite{misc-1}, for guiding therapeutic conversations and progress in MI sessions.




Most \ac{MI} chatbots produce dialogues through expert-written scripts and rules to ensure explainability and controllability~\cite{Survey-on-psychotherapy-chatbots,MI-chatbot,Zhang2020ArtificialIC,Sun2023VirtualSupport}.
However, this results in limited dialogue diversity and high costs for involving domain expertise in dialogue design.
With the advent of \ac{NLG}~\cite{nlg,gatt2018survey},
several \ac{MI} chatbots focus on rephrasing client utterances and generating \ac{MI} dialogues with templates~\cite{Almusharraf,smoking-chatbot,domin2023verve}.
Studies explored how to integrate therapeutic expertise~\cite{welivita2023boosting,li2023understanding}, such as counseling strategies, into the dialogue generation process~\cite{welivita2023boosting,tu2022misc,li2023understanding}.
\textcolor{black}{However, this is limited by reliance on domain-specific data required by NLG approaches such as pre-training~\cite{pre_training} or fine-tuning~\cite{fine_tuning}}.

The emergence of \Acp{LLM}~\cite{naveed2023comprehensive} presents new prospects for generating diverse, flexible, and engaging dialogues \textcolor{black}{in data-scarce domains.}
In addition, in-context learning with few-shot capabilities enables the integration of \ac{MI} expertise into the generation process with LLMs~\cite{Madotto,peng-etal-2020-shot}. 
Notwithstanding these promising advantages, LLM-generated dialogues face challenges in controllability and explainability to elicit behavior changes in sensitive contexts such as psychotherapy~\cite{sun2024eliciting}.
Using domain expertise effectively is the key to improving controllability and explainability in the generation process~\cite{welivita2023boosting,tu2022misc,li2023understanding}.
Inspired by the concept of Chain-of-Thoughts~\cite{cot_1,cue-cot}, we explore utilizing \acp{LLM} to predict the next therapist's MI \textit{strategies}, i.e., MI skills code~\cite{misc-1} with its definition as internal reasoning, and generate the therapist's utterance subsequently strictly following the MI strategies from reasoning. 
The present study aims to answer the following research questions: 
\begin{enumerate}[label=(\textbf{RQ\arabic*}), leftmargin=*, nosep]
    \item \textcolor{black}{How well do MI strategies align LLMs in generating MI-adherent dialogues?}
    \item \textcolor{black}{How effectively do LLMs predict MI strategies aligned with MI principles}?
\end{enumerate}

To this end, we conduct extensive experiments to assess the effectiveness of strategy-aligned MI dialogue generation, using both automatic metrics and human evaluation from MI experts and lay evaluators.
Our findings demonstrate that MI strategy can effectively instruct LLMs to generate dialogues adherent to MI principles.
\textcolor{black}{It enables the controllability and explainability of adopting LLMs to real-world MI applications such as MI chatbots for psychotherapeutic interventions.}

The contributions of this work to the current body of knowledge are three-fold: 
1) \textcolor{black}{The study is the first to investigate the use of MI strategies to align LLMs for controllable and explainable dialogue generation;}
2) We utilize both automatic and human evaluations to validate the effectiveness of such alignment;
3) \textcolor{black}{We combine theoretical analysis and empirical evidence to support the findings.}

\begin{figure*}[t!]
\centering
\includegraphics[width=0.99\textwidth]{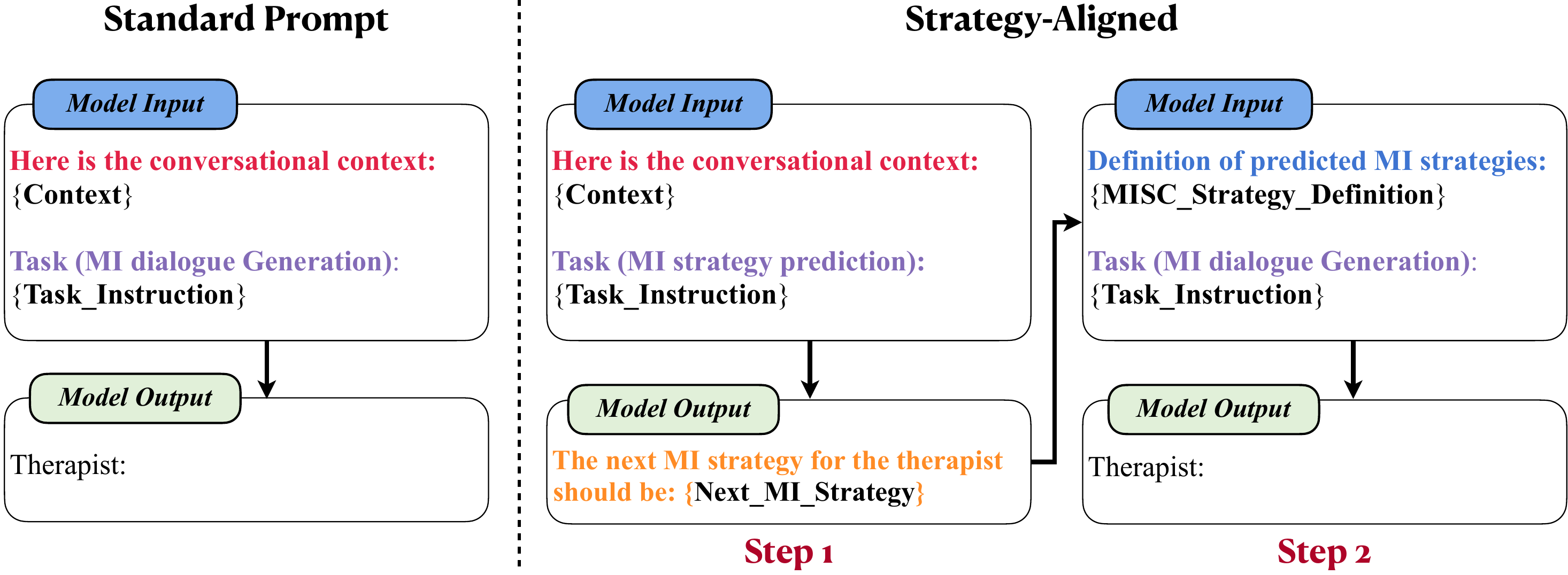}
\vspace{-0.0mm}
\caption{\done{The ``strategy-aligned'' dialogue generation with the strategy prediction as step 1 and the subsequent therapist's utterances generation as step 2 in the context of Motivational Interviewing.}}
\label{fig:prmopt-example-1}
\vspace{-1.0mm}
\end{figure*}

\section{Related Work}


\subsection{NLG in Motivational Interviewing}
Motivational Interviewing (MI) is a therapeutic counseling technique aimed at encouraging self-directed 
motivation to change behaviors~\cite{mi-1, mi-2}. 
It emphasizes empathic interaction, respects the client's autonomy, and has proved to be highly effective in motivating behavior change. As a conversational technique, it can be flexibly incorporated in various therapeutic settings. 
Experts examine MI-based interactions with strategic schemes such as the Motivational Interviewing Skill Code (MISC)~\cite{misc-1}. This coding scheme is used to assess the effectiveness of MI sessions through the quantification of essential counseling skills and adherence to MI principles. 

The role of NLG in MI has revolutionized the way digital therapeutic support is provided. 
Initially, NLG in MI was limited to replicating ongoing conversations, relying on pre-scripted templates from MI experts~\cite{Almusharraf,smoking-chatbot,welivita2023boosting}. 
The rapid advancements in LLMs now enable these models to rephrase what clients say, reflecting their words or even emotions, in ways that appear empathetic~\cite{reflective_rephrase,Rose2022,shen-etal-2020-counseling}. This development shows promising potential in enhancing client engagement and adherence to therapeutic goals.
\textcolor{black}{Despite these benefits, integrating NLG capability and applying LLMs into MI applications presents challenges, particularly in ensuring the generated content adheres to MI principles while being controllable, explainable, and free from harm.}
There are also ethical considerations, including the impact of generative AI on therapist-client relationships and the risks of LLMs generating inappropriate responses.


\subsection{\textcolor{black}{Instructed Dialogue Generation}}
The \textcolor{black}{instructed dialogue generation~\cite{InstructDial, grounded_1}} marks a significant shift in NLG, moving from focusing on linguistic fluency to incorporating specific dialogue objectives. 
This transition is evident in the progression from rule-based systems, which rely on static dialogue scripts, to generative models that adapt to dialogue contexts and specific \textcolor{black}{instructions~\cite{InstructDial}, intent~\cite{intent_jiahuan} or strategy~\cite{welivita2023boosting}, and being able to engage in ``mixed-initiative'' dialogues~\cite{tu2022misc,deng2023knowledgeenhanced}}
where models and users can both lead conversations with specific dialogue objectives.
In the realm of therapeutic dialogue generation, instructed dialogue generation efforts, such as those by~\cite{Therapeutic_Skills_Affect_Clinical,MI-Strategies,Empathetic,gao2023cab}, highlight the importance of embedding psychological and empathetic principles into response generation, aiming for alignment with therapeutic goals while maintaining dialogue engagement~\cite{Human_AI_collaboration_enables_more_empathic}.
\textcolor{black}{These works inspire our current exploration of enhancing the applications of LLMs in MI.} By aligning LLMs with specific MI strategies, we expect these can improve (i.e., strategically aligned and MI-adherent) dialogues. This approach has the dual benefit of not only controlling LLMs with MI principles but also providing more transparent generations with explicitly stated strategic objectives.
Additionally, advanced reasoning capabilities of LLMs, such as the Chain-of-Thought (CoT) concept~\cite{cot_1}, \textcolor{black}{further pave the way for strategic alignment by enabling LLMs to predict the next MI strategy as internal reasoning~\cite{mi_pred_1}.} 
Besides, in-context learning~\cite{llm_1} has emerged as a critical approach for effectively utilizing LLMs by selecting contextual prompts~\cite{cot_6}. These advancements enhance LLMs' ability to generate dialogues that are well-grounded and contextually coherent~\cite{cue-cot}.




\section{\textcolor{black}{Aligning LLMs with MI Strategy}}

\subsection{\textcolor{black}{Approach}}

Inspired by~\cite{li2023understanding}, we \textcolor{black}{propose an approach we denote ```strategy-aligned''}, to generate MI dialogues with LLMs guided by domain-specific MI strategies (i.e., the MISC~\cite{misc-1}), thereby making LLMs align to such strategies with controllability and explainability.
We employ the concept of Chain-of-Thought (CoT)~\cite{cot_1,cot_2}, enabling LLMs to internally reason the next MI strategies as the prediction based on dialogue context.
Subsequently, LLMs generate dialogues aligned with the predicted MI strategy.
For example, LLMs can generate empathetic utterances to promote therapeutic support guided by MI strategies ``Reflection''.
Figure~\ref{fig:prmopt-example-1} demonstrates the approach \textcolor{black}{``strategy-aligned'', which combines two steps: MI strategy prediction and strategy-aligned MI dialogue generation.}


\subsection{Prompt Design}
\label{prompt_design}
We design two prompting methods for our experimental purpose, including the Standard Prompt and Strategy-aligned Prompt as illustrated in Figure~\ref{fig:prmopt-example-1}. 
Detailed prompt design is attached in~\ref{appendix: prompt_template}.


\subsubsection{Standard Prompt}
For the standard prompt, we only include the dialogue context and task instruction to generate the next therapist's utterances. The objective is defined as below. 
``c'' represents the MI dialogue context; 
``u'' represents the next utterance of the therapist we expect the LLM to generate; 
``k'' is the number of dialogue sizes in the context and we choose it as 5:
\vspace{-0.0mm}
\begin{equation}
\mathcal{M} : c_{i-k, i-1} \rightarrow u_{i}
\end{equation}
\vspace{-3.0mm}


\subsubsection{Strategy-aligned Prompt}
Compared with the standard prompt, the strategy-aligned prompt is dynamic to give more specific information to LLMs. 
Specifically, we first request LLM to predict the next MISC strategy for therapists as the internal reasoning and use the reasoning output as the internal state in the subsequent MI dialogue generation process. 
Strategy-aligned prompting has three elements: 
1) the MI dialogue context;
2) the definition of MI strategies as based on the MISC~\cite{misc-1}. 
3) the LLM-predicted MI strategies of the therapist's next utterance (i.e., the type of MI strategy we expect the LLM to predict). 
We define the objective as follows, in which ``s'' stands for MI strategies; ``d'' stands for the definition of MI strategies:
\vspace{-0.0mm}
\begin{equation}
\mathcal{M} : c_{i-k, i-1}, s_{i-k, i-1}, d_{str} \rightarrow s_{i} \rightarrow u_i
\end{equation}
\vspace{-3.0mm}



\section{Experiments}

\subsection{Task Definition}
We conduct comparative experiments across benchmark LLMs using different prompting types to assess their performance in generating dialogues adherent to MI principles. 
\textcolor{black}{Specifically, we define two experimental tasks to address the research questions accordingly:
\textbf{1)} MI dialogue ``generation'' and \textbf{2)} MI strategy ``prediction''.}
The procedure of the experiments is illustrated in Figure~\ref{procedure}.



\subsection{Datasets}

Two MI datasets are used with MI strategy annotated by scheme called MISC~\cite{misc-1}. 
The first dataset is ``AnnoMI''~\cite{annomi_1,annomi_2} which has MI conversations and a single coarse-grained MI strategy per utterance. 
The second dataset is ``BiMISC''~\cite{sun2024eliciting} with MI conversations and multiple fine-grained MI strategies per utterance.
Table~\ref{tab:dataset} is an example of MI dialogues and MI strategies in these two datasets. 
Detailed MI strategies are attached in~\ref{tab:code_manual}.

\begin{table}[htbp]
\renewcommand{\arraystretch}{0.84}
\centering
\footnotesize
    \begin{tabular}{@{}lcc@{}}
    \toprule
    \multirow{2}{*}{\textbf{Therapist}} & 
    \multicolumn{2}{c}{\textbf{Strategy}} \\
    \cmidrule{2-3} 
    & 
    
    \textbf{BiMISC} & \textbf{AnnoMI} \\ \midrule
    \colorbox{pink}{That is a good example you give.}    
    & AFF & \multirow{5}{*}{RF} \\
    
    \begin{tabular}[c]{@{}l@{}}
    The sense of smoking is \\ not only motivated by the need [..] 
    \end{tabular} 
    & GI & \\
    
    \colorbox{cyan}{You say if you don't see it [...]}
    & CR & \\ \bottomrule
    \end{tabular}
    
\caption{MI strategies in BiMISC and AnnoMI datasets. 
\textbf{AnnoMI}: utterance is annotated with a single MI strategy, such as ``RF'' (reflection).
\textbf{BiMISC}: utterance is annotated with multiple strategies, like ``AFF'' (affirm), ``GI'' (give information), and ``CR'' (complex reflection).
}
\vspace{-1.0mm}
\label{tab:dataset}
\end{table}


\subsection{Benchmark LLMs}
We benchmark several prominent LLMs, focusing on LLMs renowned for their size, performance, and open-source nature. 
We select six open-sourced LLMs: \textbf{Flan-t5-xxl, Vicuna-13B, Qwen-14B,  Qwen2-7B, Llama-2-13B}, and \textbf{Llama-3-8B}.
All these open-sourced LLMs are recognized for their capability to align closely with human instructions~\cite{rlhf}, particularly in dialogue interactions. 
Additionally, we choose 
\textbf{GPT-4} as a commercial benchmark, noted for its superior performance in dialogue generation scenarios.



\subsection{Automatic Evaluation Metrics}
\label{automatic_metrics}
To objectively evaluate the quality of generations, we apply following automatic evaluation metrics. 

\begin{itemize}[nosep, leftmargin=*]


\item \textbf{BLEU \& ROUGE}~\cite{bleu,rouge} 
assesses the overlap of n-grams 
between the generation and reference in terms of precision and recall, respectively. We measure $n=1$.

\item \textbf{METEOR}~\cite{meteor} 
evaluates semantic and syntactic accuracy, including synonym and paraphrase use for linguistic precision. 

\item \textbf{BERTScore}~\cite{BERTScore} 
assesses semantic similarity by BERT embeddings, measuring contextual relevance of generations. 

\item \textbf{Entropy}~\cite{entropy} 
\done{quantifies the unpredictability and assesses the effectiveness of strategy in controlling generation. Lower entropy indicates more aligned responses.}

\item \textbf{\done{Belief}}~\cite{bayes} is 
for hypothesis testing updates posterior probabilities of generations under both hypotheses (\( H_0 \): MI strategies are effective, and \( H_1 \): they are not). The mathematic derivation is in~\ref{appendix-bayesian}.

\end{itemize}




\subsection{Human Evaluation}
\label{Human Evaluation}
We employ a two-phase human evaluation: one with MI experts to capture the alignment and adherence to MI principles and another with laypeople focusing on client-perceived quality.

\textcolor{black}{We conduct expert evaluation to assess two aspects of alignment: 
1) how effective the LLM-generated dialogues align with the MI strategies and adhere to MI principles, and 
2) how well the LLM-predicted MI strategies adhere to therapeutic MI principles.}
We select \done{100} MI dialogue contexts from datasets for evaluation.
The expert assessment involving \done{six} experts explores the nuanced effectiveness of strategy-aligned MI dialogue generation by LLMs.

The expert assessment focuses on six criteria (EC): 
(\textbf{EC1}) how effectively the MI strategy guides the generation of utterance;
(\textbf{EC2}) how independent the generated utterance is with the MI strategy; 
(\textbf{EC3}) how well the generated utterance aligns with the dialogue context; 
(\textbf{EC4}) how well the generated utterance aligns with the MI principles; 
(\textbf{EC5}) how the quality of the generated utterance compares to that of a human therapist;
(\textbf{EC6}) how well the MI strategy aligns with the dialogue context and MI principles.
The first five criteria (EC1-EC5) assess the strategic alignment of generated utterances with the MI strategy. 
The last criterion (EC6) assesses the performance of MI strategy prediction with LLMs adherent to MI principles.
Detailed assessing statements of these criteria are in Table~\ref{tab:ec_items}.

\newcommand{\ecsmall}{\fontsize{7.8pt}{7.0pt}\selectfont}
\begin{table}[!ht]
\centering
\ecsmall
\setlength{\tabcolsep}{1mm} 

\begin{tabular}{@{\extracolsep{\fill}}p{0.75cm}p{6.7cm}@{}}
\hline
Criteria & \makecell{Assessing Statement} \\ 
\toprule

    EC 1  & 
    MI strategy effectively guides therapist’s utterance generation  
    \\  
    
    EC 2  & 
    MI strategies impact the therapist’s utterance generation  
    \\  
    
    EC 3  &
    generated therapist’s utterance aligns with the dialogue context 
    \\  
    
    EC 4  & 
    generated therapist’s utterance aligns with the MI principles 
    \\  
    
    EC 5  & 
    quality of generated utterances is comparable to therapists  
    \\  

    EC 6  & 
    MI strategy aligns with dialogue context and MI principles 
    \\  
    \bottomrule
\end{tabular}
\caption{Assessing items of expert criteria (EC1-EC6).}
\label{tab:ec_items}
\vspace{-1mm}
\end{table}


Moreover, we are interested in client perceptions. 
We select \done{200} MI contexts from datasets for assessment.
\done{55} lay evaluators assess generated utterances and references from human therapists. 
\textcolor{black}{Each evaluator assesses 14 MI contexts with three criteria: appropriateness, coherence, and relevance}~\cite{eval_1,eval_2}.

\section{Outcomes}

\newcommand{\verysmall}{\fontsize{19.0pt}{20pt}\selectfont}
\begin{table*}[!ht]
\renewcommand{\arraystretch}{0.99} 
\verysmall 
\centering
\setlength\tabcolsep{8pt}
\resizebox{\textwidth}{!}{
\begin{tabular}{@{}lrrrrrrrrrrrrrrrrrrrrrrrrrr@{}}
\toprule
Model & \multicolumn{4}{c}{Length} & \multicolumn{4}{c}{BLEU$\uparrow$} & \multicolumn{4}{c}{ROUGE$\uparrow$} & \multicolumn{4}{c}{METEOR$\uparrow$} & \multicolumn{4}{c}{BERTScore$\uparrow$} & \multicolumn{4}{c}{Entropy$\downarrow$} & \multicolumn{2}{c}{Belief$\uparrow$} \\ 
\cmidrule(lr){2-5} \cmidrule(lr){6-9} \cmidrule(lr){10-13} \cmidrule(lr){14-17} \cmidrule(lr){18-21} \cmidrule(lr){22-25} \cmidrule(lr){26-27}  
MI Strategy   &     & /wo    & /w    &     &     & /wo    & /w    &     &     & /wo    & /w    &     &     & /wo    & /w   &   &  & /wo    & /w &   &  & /wo   &  w &  &  \multicolumn{2}{c}{\( H_0 \)}      \\

\midrule
\multicolumn{27}{c}{AnnoMI} \\
\midrule
Flan-T5-XXL-11B &  & 13.3 & 10.9 &  &  & 10.1 & 10.9 &  &  & 8.2 & 8.4 &  &  & 10.7   & 11.6 &  &  & 84.8 & \textbf{85.6}  &  &  & \textbf{2.9}  & \textbf{2.6} & & \multicolumn{2}{c}{0.62}  \\
Vicuna-13B &  & 40.5 & 30.1 &  &  & \textbf{14.3} & \textbf{14.7} &  &  & \textbf{12.1} & \textbf{12.3} &  &  & 17.7   & 17.3 &  &  & \textbf{85.1}  & 85.5  &  &  & 4.9  & 4.5 & & \multicolumn{2}{c}{0.65}  \\
Qwen-14B &  & 38.8 & 37.7 &  &  & 7.8 & 12.5 &  &  & 6.5 & 10.4 &  &  & 10.7  & 15.1 &  &  & 62.4  & 84.4  &  &  & 3.8  & 4.5 & & \multicolumn{2}{c}{\textbf{0.82}}  \\
Qwen-2-7B &  & 59.7 & 30.7 &  &  & 11.9 & 13.8 &  &  & 10.2 & 10.5 &  &  & 17.3  & 16.0 &  &  & 84.2  & 85.2  &  &  & 5.5  & 4.6 & & \multicolumn{2}{c}{0.64}  \\
Llama-2-13B &  & 36.2 & 44.9 &  &  &  7.5 & 14.5 &  &  & 7.7 & 11.8 &  &  & 12.0  & 18.2 &  &  &  79.8 & 84.4  &  &  & 4.2  & 5.0 & & \multicolumn{2}{c}{0.75}  \\
Llama-3-8B &  & 57.6 & 61.1 &  &  &  8.7 & 8.7 &  &  & 7.6 & 8.3 &  &  & 13.8  & 14.2 &  &  &  81.1 & 80.7  &  &  & 5.1  & 5.1 & & \multicolumn{2}{c}{0.58}  \\
GPT-4 &  & 54.3 & 23.1 &  &  & 13.6 & 14.3 &  &  & 11.2 & 11.2 &  &  & \textbf{18.7}  & \textbf{18.9} &  &  & 84.1  & 85.5  &  &  & 5.3  & 4.3 & & \multicolumn{2}{c}{0.66}  \\

\midrule
\multicolumn{27}{c}{BiMISC} \\
\midrule
Flan-T5-XXL-11B &  & 31.3 & 26.5 &  &  & 9.5 & 10.8 &  &  & 9.7 & 9.7 &  &  & 11.6  & 12.2 &  &  & 82.7  & 83.8 &  &  & \textbf{2.7}  & \textbf{2.9} & & \multicolumn{2}{c}{0.70} \\
Vicuna-13B &  & 51.9 & 38.0 &  &  & 8.4 & 12.1 &  &  & 8.0 & \textbf{10.0} &  &  & 13.1  & 16.7 &  &  & 82.7  & 84.3  &  &  & 4.9  & 4.7 & & \multicolumn{2}{c}{0.51} \\
Qwen-14B &  & 41.2 & 42.6 &  &  & 7.7 & 11.1 &  &  & 6.4 & 9.3 &  &  & 10.7  & 14.9 &  &  & 61.7  & 83.9  &  &  & 3.9  & 4.8 & & \multicolumn{2}{c}{\textbf{0.78}}  \\
Qwen-2-7B &  & 64.5 & 40.8 &  &  & 9.1 & 10.9 & &  & 7.7 & 8.3 &  &  & 14.5  & 14.2 &  &  & 82.1  & 84.1  &  &  & 5.5  & 4.8 & & \multicolumn{2}{c}{0.63}  \\
Llama-2-13B &  & 6.1 & 20.7 &  &  &  1.6 & 6.6 &  &  & 4.7 & 5.2 &  &  & 2.0  & 7.8 &  &  & 20.3  & 82.3  &  &  & 4.0  & 3.9 & & \multicolumn{2}{c}{0.74}  \\
Llama-3-8B &  & 61.2 & 61.1 &  &  & 7.2 & 7.2 &  &  & 7.1 & 7.4 &  &  & 11.5  & 11.9 &  &  & 81.2  & 81.0  &  &  & 5.1  & 5.0 & & \multicolumn{2}{c}{0.65}  \\
GPT-4 &  & 60.3 & 36.3 &  &  & \textbf{10.9} & \textbf{13.7} &  &  & \textbf{9.8} & \textbf{10.0} &  &  & \textbf{16.0}  & \textbf{16.9} &  &  & \textbf{83.4}  & \textbf{84.5}  &  &  & 5.4  & 4.6 & & \multicolumn{2}{c}{0.64} \\
\bottomrule
\end{tabular}}
\caption{Results from the automatic evaluation on two datasets with seven benchmark LLMs and two different types of prompt: standard without strategy (/wo) vs. strategy-GT with strategy (/w). Belief is the Bayesian post probabilities for hypothesis \(H_0\) that ``MI strategies are effective for MI-adherent dialogue generation''.}
\vspace{-0.0mm}
\label{tab:automatic-eval-res}
\end{table*}
\begin{figure*}[!htbp]
\centering
    \includegraphics[height=4.694cm]{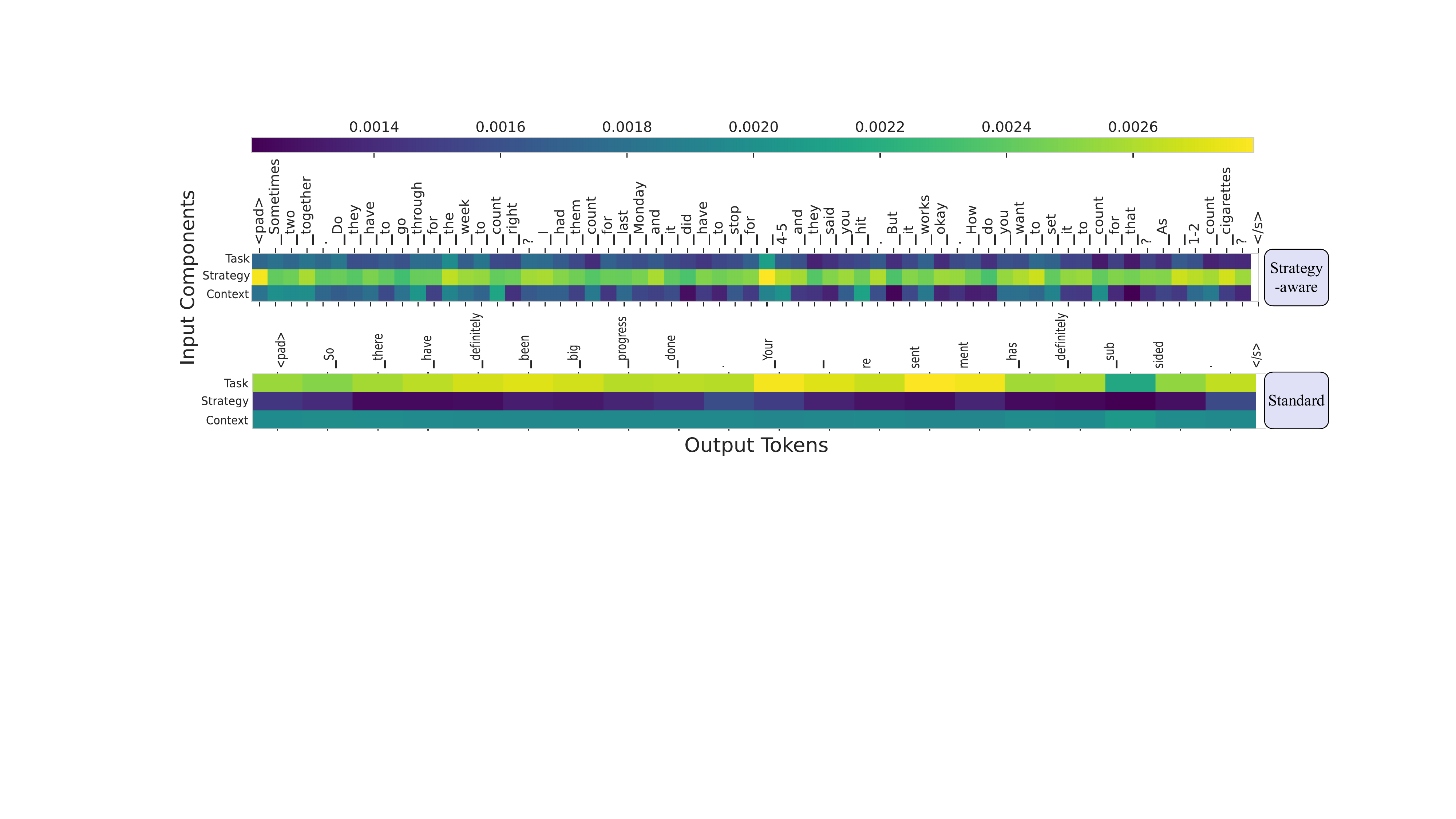} 
\caption{Comparison of attention score distributions from LLM (i.e., Flan-T5 in \ref{appendix: llms} with Encoder-Decoder architecture, last layer, and averaged across all heads) for dialogue generation, with (``strategy-aligned'') and without (``standard'') the MI strategy. The input conversational context and strategy match the 'Case Study' example for consistency. Attention to input tokens is aggregated into three prompting components for better comparison.}
\vspace{-1.0mm}
\label{attention}
\end{figure*}


\subsection{Empirical Analysis on Automatic Metrics}

Table~\ref{tab:automatic-eval-res} demonstrates the standard prompt yields the lowest scores in automatic metrics. 
This outcome shows that strategy-aligned generation with MI strategy can effectively instruct LLMs to generate dialogue following specific MI principles.
\textcolor{black}{From the model perspective, the commercial GPT-4 model consistently achieves the highest scores across metrics.} However, notable is the performance of open-sourced LLMs Flan-T5, Vicuna-13B and Qwen2, which closely rival that of GPT-4. This highlights the significant advancements in open-sourced LLMs for MI dialogue generation.





To comprehensively evaluate the effectiveness of MI strategies for MI-adherent dialogue generation, we applied Bayesian inference~\cite{bayes}, which allows to update the belief in a hypothesis based on multiple automatic metrics listed in Table~\ref{tab:automatic-eval-res} (i.e., BLEU, ROUGE, BERTScore, and Entropy). We calculated the likelihood of each generation for two hypotheses:
\( H_0 \) (MI strategy is effective for dialogue generation) and 
\( H_1 \) (MI strategy is not effective).
\textcolor{black}{Our findings show that all LLMs exhibited a higher probability for \( H_0 \), indicating that MI strategies are generally effective for guiding dialogue generation. Notably, open-source models Qwen 14B and Llama2-13B demonstrated the highest belief in \( H_0 \), surpassing GPT-4. This highlights the potential of open-source LLMs in strategic alignment tasks, offering promising alternatives to proprietary models.}


Moreover, to understand how LLMs utilize MI strategies in dialogue generation, Figure~\ref{attention} visualizes the attention distribution~\cite{visualization_attention} of LLM generations with and without MI strategy. The attention distribution for the strategy-aligned generation shows a significantly denser focus on the MI strategy compared to the standard prompting components (i.e., task instruction and dialogue context only). \textcolor{black}{This emphasizes LLM's consideration of MI strategies in strategy-aligned dialogue generation.}




\subsection{Aligning LLM with MI Strategy for Dialogue Generation}
\begin{figure*}[!htbp]
\centering
\includegraphics[width=0.99\textwidth]{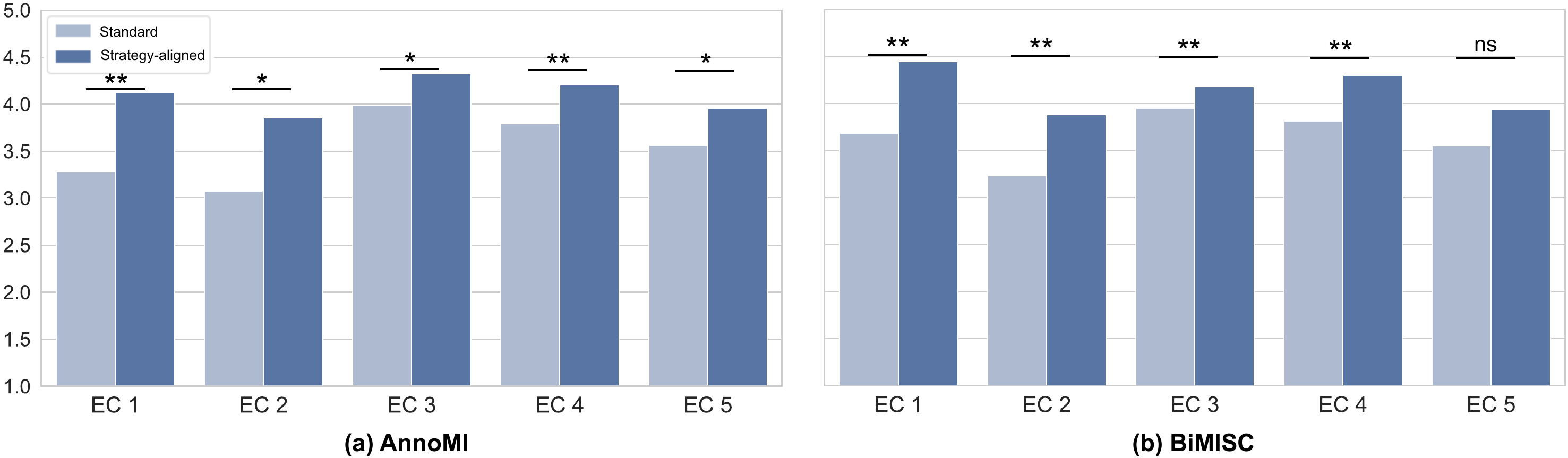}
\vspace{-0.0mm}
    \caption{
     Experts evaluation on two datasets based on assessing criteria (``EC1-EC5'') in Table~\ref{tab:ec_items}.
     It assesses the alignment between MI strategy and utterances generated by ``Standard'' and ``Strategy-aligned'' prompts.
     The y-axis denotes the average ratings ranging from 1 (Strongly disagree) to 5 (Strongly agree). (**$p$<.01, *$p$<.05, ``ns'' $p$<.1)
     }
\vspace{-1.0mm}
\label{expert-eval-alignment}
\end{figure*}

In expert evaluations, MI experts assess the alignment and quality of LLM-generated MI dialogues with \textcolor{black}{either ``Standard'' or ``Strategy-aligned'' prompting methods} in Section~\ref{prompt_design}, focusing on determining how well the generated utterances align with the MI strategies and dialogue context.
Figure~\ref{expert-eval-alignment} and paired-samples t-test~\cite{t-test} show that ``Strategy-aligned'' is significantly more effective in guiding the generation of utterances than standard prompt by criteria EC1 on both datasets ($p<.01$) and the generated utterances are significantly more dependent on MI strategy (EC2) on AnnoMI ($p<.05$) and BiMISC ($p<.01$), proving the effectiveness of MI strategy in aligning dialogue generation with MI principles. 
\textcolor{black}{Further analyses revealed that the quality of generated utterances with ``strategy-aligned'' are significantly higher than the generations with the standard prompt (EC3 \& EC4) ($p<.01$): both prompting approaches achieving above-average scores compared to human therapist's utterances (EC5), indicating the potential of LLMs in generating therapeutic dialogues in MI.}
Thus, expert evaluation solves our first research question that MI strategy can guide LLMs to generate dialogues that are strictly aligned with MI principles and are comparable to those of human therapists.


\textcolor{black}{
Lay people's evaluations offer valuable insights from the client's perspective, complementing expert assessments. 
As shown in Table~\ref{laypeople-eval}, Vicuna model using "strategy-aligned" prompts scored significantly higher than both standard prompt and reference utterances across all three assessment dimensions: ``Appropriateness'', ``Coherence'', and ``Empathy'' with $p<.01$ on both datasets, confirmed by post hoc Tukey HSD tests~\cite{tukey_test}. 
This underscores the effectiveness of the MI strategy in aligning Vicuna with MI principles.
However, GPT-4 showed a different trend, with a significant improvement in ``Empathy'' ($p<.05$) of the standard prompt, but not in ``Appropriateness'' ($p=.07$) or ``Coherence'' ($p=.19$). 
This suggests that while experts confirm the effectiveness of MI strategy aligning GPT-4 with MI principles (results from Figure~\ref{expert-eval-alignment}), it might affect the empathetic conversational nature from the client's perspective~\cite{Machine_and_Human_Understanding_of_Empathy}.
}
A further ``Case Study'' provides concrete examples demonstrating a nuanced balance between MI strategic alignment and client perceptions.



\vspace{0.0mm}


\newcommand{\laysmall}{\fontsize{8.80pt}{5.6pt}\selectfont}
\begin{table}[!ht]
\renewcommand{\arraystretch}{1.0} 
\laysmall 
\setlength\tabcolsep{2.90pt}
\centering
\begin{tabular}{lrrrrrrrrrr}
\toprule

Dataset & \multicolumn{5}{c}{AnnoMI} & \multicolumn{5}{c}{BiMISC} \\ 

\cmidrule(lr){2-6} \cmidrule(lr){7-11}

Model/Prompt   &     & App  & Coh  & Emp   &    &     & App  & Coh   & Emp  & \\ 

\midrule
Reference &  & 2.77 & 2.93 & 2.92 &  &  & 2.71 & 2.65 & 2.87 & \\

Vicuna-Std &  & 3.75 & 3.69 & 3.70 &  &  & 3.62 & 3.57 & 3.59 & \\

Vicuna-Aligned &  & 3.77 & 3.75 & 3.77 &  &  & 3.81 & 3.79 & 3.85 & \\

GPT-4-Std &  & 3.91 & 3.88 & 3.94 &  &  & 4.00 & 3.94 & 3.99 & \\

GPT-4-Aligned &  & 3.71 & 3.72 & 3.68 &  &  & 3.80 & 3.79 & 3.87 & \\

\bottomrule
\end{tabular}
\vspace{-0mm}
\caption{Mean of laypeople assessment on generated utterances by either standard (-Std) or ``strategy-aligned'' (-Aligned) prompts from Vicuna and GPT-4 models, and reference utterances from therapists. ``App'': Appropriateness; ``Coh'': Coherence; ``Emp'': Empathy.}
\vspace{-1.2mm}
\label{laypeople-eval}
\end{table}

\subsection{\textcolor{black}{Predicting MI Strategy as Reasoning}}

We explore how effectively LLMs predict MI strategies aligned with MI principles to address RQ2.
Building on work~\cite{mi_pred_1}, we first benchmark LLMs for the MI strategy prediction task. 
As shown in Table~\ref{tab:automatic-cos}, GPT-4 achieves the highest accuracy. \textcolor{black}{Specifically, GPT-4 is the pre-trained model with zero-shot setup, while GPT-4o (FT) is the model fine-tuned on two datasets. 
Fine-tuning significantly improves LLM performance for MI strategy prediction.}
Additionally, the accuracy drops in BiMISC with multiple fine-grained strategies compared to AnnoMI with a single coarse-grained strategy (as shown in Table~\ref{tab:dataset}), indicating higher complexity of multi-label prediction in MI context. 

\begin{table}[!ht]
\renewcommand{\arraystretch}{0.760} 
\footnotesize
\setlength\tabcolsep{6.0pt}
\centering
\begin{tabular}{lrrrrrrrr}
\toprule

Dataset & \multicolumn{4}{c}{AnnoMI} & \multicolumn{4}{c}{BiMISC} \\ 

\cmidrule(lr){2-5} \cmidrule(lr){6-9}

Metric   &     & Acc    & F1    &    &     & Acc    & F1   & \\ 

\midrule
Flan-T5 &  & 46.2 & 77.6 &  &  & 19.1 & 17.4 \\

Vicuna-13B &  & 44.7 & 76.2 &  &  & 10.5 & 18.8 \\

GPT-4 &  & \textbf{50.0} & \textbf{78.9} &  &  & \textbf{33.6} & \textbf{27.9} \\

GPT-4o (FT) &  & \textbf{63.6} & \textbf{81.4} &  &  & \textbf{47.2} & \textbf{36.9} \\

\bottomrule
\end{tabular}
\vspace{-0.0mm}
\caption{The next MI strategy prediction for single-strategy in AnnoMI vs. multiple-strategy in BiMISC.}
\vspace{-1.6mm}
\label{tab:automatic-cos}
\end{table}


\begin{figure}[!ht]
\centering   
\includegraphics[width=0.475\textwidth]{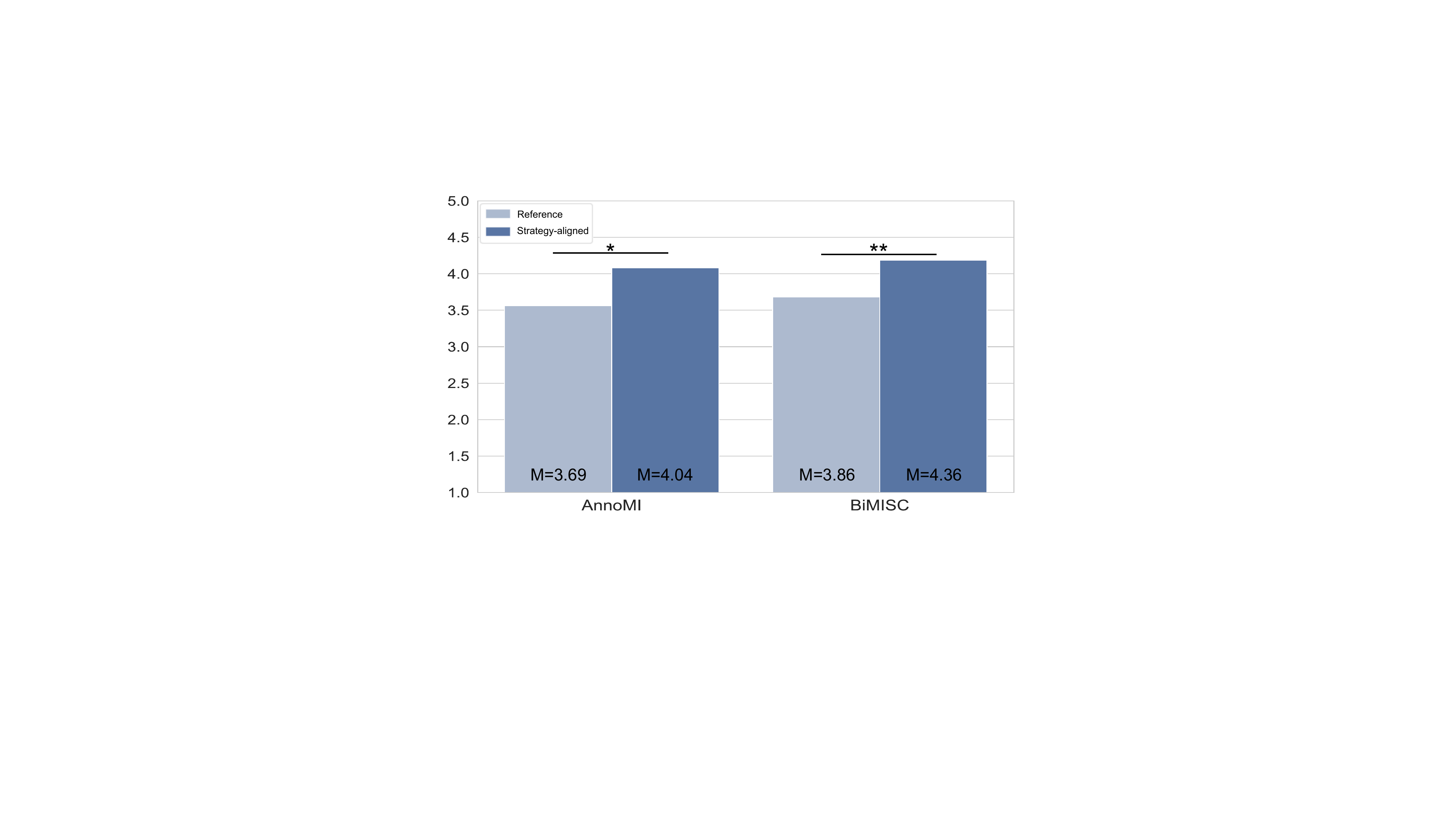} 
\vspace{-0.6mm}    
\caption{Experts assess the MI strategy prediction by GPT-4 using criteria (EC6): ``MI strategy aligns with dialogue context and MI principles''.
(**$p$<.01, *$p$<.05)}
\vspace{-1.20mm}
\label{expert-eval-planning}
\end{figure}

\textcolor{black}{Although the performance of prediction in Table~\ref{tab:automatic-cos} is modest and the strategy is incorrectly predicted compared to ground truth, the LLM-predicted MI strategies could still be appropriately aligned with context and MI principle, because the MI strategies in MI sessions are usually not uniquely appropriate. 
``Case Study'' further gives an example demonstrating such nature of MI.
Therefore, we conduct the expert evaluation to assess the alignment of the predicted MI strategy with dialogue context and MI principles.
According to expert criteria (EC6), Figure~\ref{expert-eval-planning} shows that  ``strategy-aligned'' demonstrates significantly superior contextual alignment and adherence to MI principles compared to ``standard'' (i.e., the LLM-generated utterances by Standard prompt and the ground truth MI strategy annotated in the datasets) (with \(p<.05\) in AnnoMI and \(p=.001\) in BiMISC) by paired-samples t-test. 
This highlights the significant effectiveness of LLMs for MI strategy prediction.}

\begin{figure*}[!htbp]
    \centering
    \includegraphics[width=0.988\textwidth]{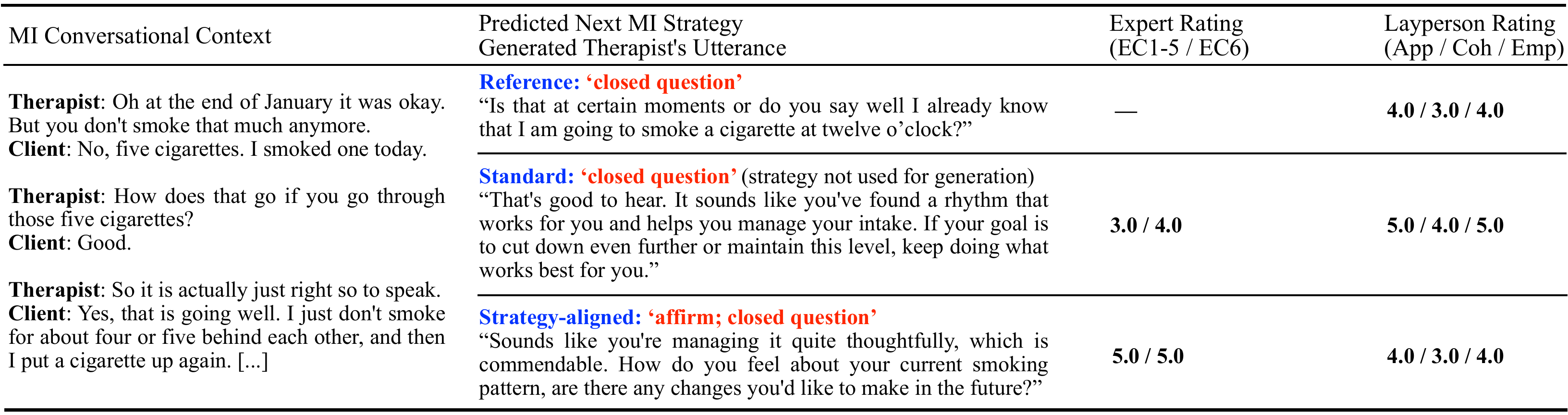} 
    \vspace{-2.0mm}
    \caption{An example of MI dialogue generated by two prompting methods as well as a reference from a human therapist, assessed by MI experts and laypeople.}
    \label{case_study}
    \vspace{-3.2mm}
\end{figure*}

\section{Case Study}

This case study aims to shed light on:
1) LLM-predicted strategies can still align with MI principles and context, although prediction differs from ground-truth strategy;
2) balance between therapeutic strategic alignment and client perception.

Figure~\ref{case_study} shows LLM-predicted MI strategies ``affirm; closed question'' in ``strategy-aligned'' receives a higher alignment expert score (EC6) (5.0) than ground-truth strategy (3.0) as only ``closed question'', indicating more effectiveness and empathy, and illustrating the potential of LLM for predicting MI strategy.
\textcolor{black}{Therefore, expert evaluations in Figure~\ref{expert-eval-planning} argue that while the prediction accuracy in Table~\ref{tab:automatic-cos} is modest,} some inaccurately predicted strategies are still appropriate and adhere to MI principles within the specific dialogue context, due to the non-uniqueness nature in MI context.

\textcolor{black}{Moreover, experts give higher scores for alignment between generated utterances and strategy (EC1-5) in ``strategy-aligned'' compared to standard prompt, indicating that strategy-aligned generations are more aligned with MI principles which are ``affirmative'' and ``questional'' for controlling the dialogue flow. 
However, lay evaluators rate generations of standard prompt higher, especially in dimension of ``empathy''. The utterance ``How do you feel about your current smoking
pattern, are there any changes you'd like to make in the future?'' generated by ``strategy-aligned'' is strictly aligned with MI strategy ``closed question'', whereas the standard prompt generates ``That's good to hear. [...], every small step is progress.'' reflects more empathy as it is MI strategy of ``reflection'' instead of the ``closed question'', which can not lead dialogue to next topic in MI session.}
Laypeople may perceive dialogues incorporating strict MI strategies as rigid or lacking emotional nuances, suggesting a gap between strategic alignment and client perceptions. This underscores the complexity of translating MI adherence into dialogues perceived as empathetic, highlighting the need for LLMs to balance MI strategies with greater naturalness in psychotherapeutic conversations.


\section{Discussion}

\subsection{Strategy-Aligned MI Dialogue Generation}


Current MI applications rely heavily on expert pre-scripted dialogues, which are time and resource-intensive and require significant domain expertise for conversational design. These pre-scripted dialogues often lack dialogue diversity, which is crucial in MI.
LLMs present a promising solution by generating diverse and coherent dialogues, introducing a greater variety of responses, and enhancing the adaptive nature of MI dialogues, in addition to reducing dependency on experts.

The uncontrollable nature of LLMs, however, poses risks in sensitive areas like psychotherapy. LLM-generated dialogues can deviate from MI principles and therapeutic goals, potentially leading to inappropriate outputs. This is where strategic alignment with MI principles becomes critical.
Ensuring LLM-generated content adheres to MI guidelines safeguards against deviations that could undermine the therapeutic process. Integrating MI strategies into dialogue generation helps LLMs produce utterances that are both relevant and consistent with MI principles.
Additionally, strategic alignment provides explainability in MI dialogue generation, which is essential for applying LLMs in sensitive contexts like psychotherapy. It enhances the safety and perceived trustworthiness of LLM-generated outputs, making the integration of LLMs into MI applications more acceptable.

Extensive evaluations in this work indicate the great potential of strategic alignment in generating MI dialogue with controllability and explainability by ensuring that LLMs follow MI strategies, bridging the gap between automated dialogue generation and therapeutic nuances of effective MI. 


\subsection{Challenges of Applying LLMs in MI}

Applying LLMs in MI presents distinct challenges. 
This work indicates that strategy can effectively guide LLM-generated dialogues adherent to MI principles. While strategy-aligned utterances are technically precise within therapeutic goals, they may lack naturalness and empathy, which laypeople prioritize~\cite{Machine_and_Human_Understanding_of_Empathy}.
This highlights the need for balances between empathetically engaging utterances and those aligned with MI principles. 
Achieving such balance is crucial for the success of LLM-assisted psychotherapeutic conversation tools, which must meet professional standards while resonating with people, ensuring both strategic accuracy and genuine human connection.

Moreover, the way LLM-generated utterances influence dialogue flow is pivotal. For instance, the generated question utterances could alter the expected course of therapist-led interactions, affecting the ongoing dialogue flow in MI. MI strategy prediction by LLMs provides the potential to ensure that dialogues are contextually relevant and adhere to MI principles.
It makes the generative and mixed-initiative systems~\cite{tu2022misc} more controllable and aligned with domain expertise. 
However, inappropriately predicted strategies could lead dialogues in unintended directions, detracting from the therapeutic goals. Enhancing LLMs' ability to accurately understand and predict MI strategies at each reasoning step is key to ensuring dialogues remain on course.
Future advancements in LLM reasoning enhanced by domain adaption and further fine-tuning are crucial to overcome these challenges and maximizing the potential of LLM-generated therapeutic dialogues. 




\section{Conclusion}

This work tackles the challenge of utilizing LLMs in the sensitive domain of psychotherapy, confronted by the uncontrollable nature of LLMs. 
\textcolor{black}{We rethink the way to align LLMs with MI strategy for safer MI dialogue generation.}
We conduct extensive experiments and analyses with automatic and human evaluations to validate that the MI strategy can effectively align LLMs to generate dialogues adherent to MI principles. The findings also highlight the need for balancing strategic alignment with empathetic engagement in psychotherapeutic interactions.
We provide a controllable and explainable solution for MI dialogue generation by LLMs, setting a foundation for future research to enhance the efficacy of LLMs in psychotherapy.


\section*{Limitations}


While this work provides valuable insights, several limitations should be acknowledged.

First, the generalizability of our findings is limited by the specific data and demographics used. Expanding the dataset to include a more diverse range of MI scenarios and client demographics could improve the generalizability of the findings.

Second, while ``strategy-aligned'' offers the potential to guide LLMs in MI dialogue generation, it may not fully capture the dynamic nature of human therapeutic communication and emotional nuances of client and therapist interaction. Enhancing the models with advanced techniques like domain adaption, fine-tuning, or grounding with domain-specific knowledge bases to better capture these dynamic and emotional nuances could help.


Third, the reliance on subjective human evaluations and traditional automatic metrics might not capture the full mental resonance and therapeutic effectiveness of the dialogues. To develop and incorporate more nuanced and comprehensive evaluation metrics could help.

Fourth, while the study assesses strategic alignment with MI principles, it does not measure the impact of generated dialogues on actual therapeutic outcomes like client motivation or behavior change. The practical application of LLM-empowered MI in real-world settings remains untested, and their effectiveness in live sessions needs empirical validation. Additionally, deploying LLMs in therapeutic contexts raises further ethical concerns, including handling sensitive information and potential biases in the generated content. Future research could conduct longitudinal studies to evaluate long-term effects, pilot studies or controlled trials to test real-world effectiveness with comprehensive ethical considerations.


Future work will focus on refining LLM capabilities to capture nuanced human conversational interactions and conducting empirical studies to validate the effectiveness and ethical deployment of LLMs in live therapeutic settings.




\section*{Ethical Statement}
This research adheres to rigorous ethical standards to ensure the responsible use of AI and the protection of participant data. Large Language Models (LLMs) were carefully monitored to prevent harmful or biased content, particularly in the sensitive context of MI. Informed consent was obtained from all human evaluators, ensuring their anonymity and right to withdraw at any time without consequence (see~\ref{consent} for details). 
Ethical considerations included addressing potential risks, such as handling sensitive client information and mitigating biases in generated content. This work complies with institutional ethical requirements, and future researchers are expected to uphold these standards to ensure the responsible use of data and the advancement of knowledge in the field.


\subsection*{Data Anonymization and Privacy}
Data privacy is a priority. 
Part of the data utilized in this work originates from MI counseling sessions and thus contains sensitive information. 
To protect the privacy of the individuals involved, we implement rigorous data anonymization procedures for the human-involved evaluation. 
All identifiable information, including names, addresses, and specific personal details, are meticulously removed to ensure confidentiality and anonymity.
Recognizing the potential implications of AI technology in therapeutic settings, we advocate for ongoing dialogue and collaboration between AI experts, ethicists, and psychotherapeutic professionals to guide the future development of LLM-assisted therapeutic systems and applications.

\subsection*{Use of AI Tools}
We only use the AI tools (i.e., ChatGPT and Grammarly) for checking the grammatical errors.

\subsection*{Acknowledgment}
This work is funded by the European Commission in the Horizon H2020 scheme, awarded to Jos A.Bosch (TIMELY Grant agreement ID: 101017424). 
Additionally, Zhuying Li is supported by the National Natural Science Foundation of China (No. 62302094) 
Pengjie Ren is supported by the Natural Science Foundation of China (No. 62472261, 62102234), the Key R\&D Program of Shandong Province (grant: 2024CXGC010108)
All content represents the opinion of the authors, which is not necessarily shared or endorsed by their respective employers and/or sponsors.



\newpage
\bibliography{main}



\clearpage

\clearpage
\onecolumn

\begin{appendices}
\renewcommand{\thesection}{Appendix A}
\section{Details of LLMs in the experiments}
\label{appendix: llms}
\end{appendices}

\setcounter{table}{0} 

We benchmark several prominent LLMs to evaluate their performance, focusing on LLMs renowned for their size, performance, and open-source nature. 
We select six open-sourced LLMs: 
\texttt{Flan-T5-XXL}~\footnote{\url{https://huggingface.co/google/flan-t5-xxl}\label{ft:flan-t5}}, 
\texttt{Vicuna-13B}~\footnote{\url{https://huggingface.co/lmsys/vicuna-13b-v1.5}\label{ft:vicuna}}, 
\texttt{Qwen-14B-Chat}~\footnote{\url{https://huggingface.co/Qwen/Qwen-14B-Chat}\label{ft:qwen}}, 
\texttt{Qwen2-7B-Instruct}~\footnote{\url{https://huggingface.co/Qwen/Qwen2-7B-Instruct}\label{ft:qwen2}}, 
\texttt{Llama-2}~\footnote{\url{https://huggingface.co/meta-llama/Llama-2-13b-hf}\label{ft:llama2}}, and
\texttt{Llama-3}~\footnote{\url{https://huggingface.co/meta-llama/Meta-Llama-3-8B}\label{ft:llama3}}.
All these open-sourced LLMs are recognized for their capability to align closely with human instructions~\cite{rlhf}, particularly in dialogue interactions. 
We use the ``Transformers'' package from the HuggingFace~\footnote{\url{https://huggingface.co/docs/transformers/en/main_classes/text_generation}} to do generations by all these open-sourced models. 

Additionally, we choose 
\texttt{GPT-4}~\footnote{\url{https://platform.openai.com/docs/models/gpt-4-and-gpt-4-turbo}\label{ft:gpt4}} 
as the commercial benchmark, noted for its superior performance in various NLP tasks and especially in the dialogue scenarios. 
We use \texttt{openai} Python library and \texttt{requests} library to send requests to the OpenAI API~\footnote{\url{https://platform.openai.com/docs/overview}} and to do the generations with GPT-4. 

The models were used in compliance with their respective licenses and terms at the time of the study. 
\clearpage

\begin{appendices}
\renewcommand{\thesection}{Appendix B}
\section{Bayesian inference to validate the effectiveness of MI strategy}
\label{appendix-bayesian}
\end{appendices}

\subsection*{Introduction}

Bayesian inference provides a way for us to evaluate the effectiveness of the Motivational Interviewing (MI) strategy in guiding the LLM generations. It updates the beliefs about hypotheses given new evidence (i.e., new generation in our case). More importantly, this approach allows us to evaluate the overall effectiveness of the MI strategy based on a set of automatic metrics as discussed in Section~\ref{automatic_metrics}.

\vspace{4mm}
\subsection*{Hypotheses}

\begin{itemize}
  \item \(H_0\): MI strategy can effectively guide LLM for MI-adherent dialogue generation.
  \item \(H_1\): MI strategy can NOT effectively guide LLM for MI-adherent dialogue generation.
\end{itemize}

\vspace{4mm}
\subsection*{Step 1: Initial Setup}

\textbf{Initial Data Collection}: \newline
Collect all the generated dialogues and compute the mean (\(\mu\)) and standard deviation (\(\sigma\)) for BLEU, ROUGE, METEOR, BERTScore, and Entropy for both dialogues generated with (``strategy-GT'') and without (``standard'') the MI strategy. \newline \\
\textbf{Initial Statistics}: \\
Mean (\(\mu\)) and Standard Deviation (\(\sigma\)) Value for generations with MI strategy (``strategy-GT''):
\[
\begin{aligned}
&\mu_{\text{BLEU,with}}, \mu_{\text{ROUGE,with}}, \mu_{\text{METEOR,with}}, \mu_{\text{BERTScore,with}}, \mu_{\text{Entropy,with}}\\
&\sigma_{\text{BLEU,with}}, \sigma_{\text{ROUGE,with}}, \sigma_{\text{METEOR,with}}, \sigma_{\text{BERTScore,with}}, \sigma_{\text{Entropy,with}} \\
\end{aligned}
\]
Mean (\(\mu\)) and Standard Deviation (\(\sigma\)) for generations without MI strategy (``standard''):
\[
\begin{aligned}
&\mu_{\text{BLEU,without}}, \mu_{\text{ROUGE,without}}, \mu_{\text{METEOR,without}}, \mu_{\text{BERTScore,without}}, \mu_{\text{Entropy,without}}\\
&\sigma_{\text{BLEU,without}}, \sigma_{\text{ROUGE,without}}, \sigma_{\text{METEOR,without}}, \sigma_{\text{BERTScore,without}}, \sigma_{\text{Entropy,without}} \\
\end{aligned}
\]

\vspace{4mm}
\subsection*{Step 2: Bayesian Calculation}

\textbf{Prior Probabilities}: \newline
Initially, set \(P(H_0) = P(H_1) = 0.5\), reflecting no prior preference for either hypothesis. \newline \\
\textbf{Likelihood Calculation}: \newline
For automatic evaluation metrics of all observed generations \(D\):
\[
P(D|H_0) = \prod_{i=1}^{N} \prod_{\text{metric}} \frac{1}{\sqrt{2\pi\sigma_{\text{metric,with}}^2}} \exp\left( -\frac{(D_{i,\text{metric}} - \mu_{\text{metric,with}})^2}{2\sigma_{\text{metric,with}}^2} \right)
\]
\[
P(D|H_1) = \prod_{i=1}^{N} \prod_{\text{metric}} \frac{1}{\sqrt{2\pi\sigma_{\text{metric,without}}^2}} \exp\left( -\frac{(D_{i,\text{metric}} - \mu_{\text{metric,without}})^2}{2\sigma_{\text{metric,without}}^2} \right)
\]
\newline \\
\textbf{Posterior Probability Calculation}: \newline 
Using Bayes' Theorem, calculate the posterior probabilities:
\[
P(H_0|D) = \frac{P(D|H_0) \cdot P(H_0)}{P(D)}
\]
\[
P(D) = P(D|H_0) \cdot P(H_0) + P(D|H_1) \cdot P(H_1)
\]

The posterior for \(H_1\) is similarly calculated:
\[
P(H_1|D) = \frac{P(D|H_1) \cdot P(H_1)}{P(D)}
\]

\vspace{4mm}
\subsection*{Step 3: Final Bayesian Belief Update}

The final belief in the effectiveness of the MI strategy is the \textbf{posterior probability} \(P(H_0|D)\) after considering all generations from LLM in one condition. It provides a comprehensive assessment of the effectiveness of MI strategy-aware dialogue generation using Bayesian inference, grounded in empirical generated dialogue data from the experiment. The automatic metrics (BLEU, ROUGE, METEOR, BERTScore, and Entropy) provide a robust assessment of dialogue quality.
This method ensures that the belief in the effectiveness of the MI strategy is based on a thorough analysis of all empirical data, resulting in a comprehensive calculation.
\[
\text{Final Belief for \(H_0\)} = P(H_0|D)
\]  
\newline
\subsection*{Example Calculation}

Considering we have the following metrics for dialogues generated with and without the MI strategy:

\[
\begin{aligned}
\mu_{\text{BLEU,with}} &= 3, \quad \sigma_{\text{BLEU,with}} = 0.5 \\
\mu_{\text{BLEU,without}} &= 2.5, \quad \sigma_{\text{BLEU,without}} = 0.5 \\
\mu_{\text{ROUGE,with}} &= 5, \quad \sigma_{\text{ROUGE,with}} = 1 \\
\mu_{\text{ROUGE,without}} &= 4.5, \quad \sigma_{\text{ROUGE,without}} = 1 \\
&\vdots
\end{aligned}
\]

For a new generation of MI dialogue with \( \text{BLEU} = 3.2, \text{ROUGE} = 5.5, \ldots \)

We can calculate the likelihoods \(P(D|H_0)\) and \(P(D|H_1)\):

\[
P(D|H_0) = \left(\frac{1}{\sqrt{2\pi \cdot 0.5^2}} \exp\left( -\frac{(3.2 - 3)^2}{2 \cdot 0.5^2} \right)\right) \cdot \left(\frac{1}{\sqrt{2\pi \cdot 1^2}} \exp\left( -\frac{(5.5 - 5)^2}{2 \cdot 1^2} \right)\right) \cdot \ldots
\]

\[
P(D|H_1) = \left(\frac{1}{\sqrt{2\pi \cdot 0.5^2}} \exp\left( -\frac{(3.2 - 2.5)^2}{2 \cdot 0.5^2} \right)\right) \cdot \left(\frac{1}{\sqrt{2\pi \cdot 1^2}} \exp\left( -\frac{(5.5 - 4.5)^2}{2 \cdot 1^2} \right)\right) \cdot \ldots
\]

Then, using these likelihoods, compute the posterior probabilities.

\[
P(D) = P(D|H_0) \cdot 0.5 + P(D|H_1) \cdot 0.5
\]

\[
P(H_0|D) = \frac{P(D|H_0) \cdot 0.5}{P(D)}
\]

\[
P(H_1|D) = \frac{P(D|H_1) \cdot 0.5}{P(D)}
\]

This calculation yields a final belief in the effectiveness of MI strategy in guiding LLM for dialogue generation based on empirical data from experiments considering chosen automatic evaluation metrics.


\clearpage
\onecolumn

\begin{appendices}
\renewcommand{\thesection}{Appendix C}
\section{\done{Example of the prompt template}}
\label{appendix: prompt_template}
\end{appendices}

\setcounter{table}{0} 

\begin{table}[ht!]
\centering
\footnotesize
\renewcommand{\arraystretch}{2.0}
\begin{tabularx}{\textwidth}{>{\raggedright\arraybackslash}p{0.30\textwidth} >
{\raggedright\arraybackslash}X} 
\hline
\multicolumn{1}{l}{\textbf{Component}} & \multicolumn{1}{l}{\textbf{Content}} \\
\hline

\textbf{Conversational Context} & Conversational context:\newline
\textbf{[We give historical conversations]} \newline \newline
Therapist: Yes, those were not really your moments, they were not really your smoking moments, that was a bit literally and figuratively, especially at the end of the day.\newline
[...] \newline
Client: Yes.\newline
Therapist: Yes, okay, so you say I am actually satisfied with the current state of affairs and ...\newline
Client: Yes I, I already said that, I like that with losing weight, I have a striving that I am between 85 and 90, that I still want to throw smoking out all the way, it is better anyway And cheaper.
\\ \hline

\textbf{For ``Strategy-aligned'' Only} &  The next MI strategy is: the next MISC code(s) and the definition of the MISC code(s) as well as the MISC examples (optional). 
\\ \cline{1-2}

\textbf{Next MISC Code(s) (optional)} &  The next MISC code(s) for therapist should be: \newline
\textbf{[We give the next MI strategy from the prediction]} \newline \newline 
The next MISC code(s) for the therapist should be: [the Next MI code(s)] 
\\ \cline{2-2}

\textbf{MISC Definition (optional)} &  The definition of the next MISC code(s) for therapist: \newline
\textbf{[We give the definition of MISC according to the MISC code(s)]} \newline \newline
'reflection': reflection is a statement made by the therapist that captures and mirrors back the essence of what the client has said or expressed. [...]  \newline \newline
'question': question is made by the therapist to gain more clarity or to explore the client's perspective, feelings, thoughts, or experiences. [...] \newline \newline
'therapist\_input': therapist\_input is any other therapist utterance that is not codable as 'question' or 'reflection'. [...] 
\\ \cline{2-2}

\textbf{MISC Examples (optional)} & Examples of each code in MISC: \newline
\textbf{[We can optionally give examples of MISC code(s) according to the next MISC code(s)} \newline \newline
'reflection': \newline
Example 1:\newline 
Client: 'I'm scared of the consequences if I don't stop smoking.'\newline
Therapist: 'You're expressing fear about the potential effects of continued smoking.' [...] \newline\newline
'question':\newline
Example 1:\newline
Client: 'I think I need to stop smoking.' \newline
Therapist: 'Have you tried quitting before?' [...] 
\\ \hline
\textbf{Task Instruction} & Task:\newline
\textbf{[We give instruction to explain the dialogue generation task]} \newline \newline
Given the conversational context, please generate the next therapist's utterance that strictly follows the next MISC strategy and its definition. Please only generate an utterance and do not generate question statement if the next MISC strategy is neither ``open question'' nor ``closed question''.\newline

The next therapist's utterance is: 
\\ \hline

\end{tabularx}
\caption{An example of the prompt template used in the experiments for the ``standard'' or ``strategy-aligned'' MI dialogue generation.}
\end{table}
\clearpage
\onecolumn

\begin{appendices}
\renewcommand{\thesection}{Appendix D}
\section{\done{MI strategy: MISC codes and their definitions}}
\label{tab:code_manual}
\end{appendices}

\begin{table*}[ht!]
\scriptsize
\renewcommand{\arraystretch}{1.40}
\begin{tabularx}{\textwidth}
{>{\raggedright\arraybackslash}p{0.18\textwidth} >{\raggedright\arraybackslash}p{0.46\textwidth} >{\raggedright\arraybackslash}X}
\toprule
\textbf{\begin{tabular}[c]{@{}l@{}}Therapist Code\end{tabular}} & \textbf{Description (abbreviated version)} & \textbf{Example} \\

\midrule
Question (QUS) & Asking questions for a wide range of answers. & What are your main goals for our sessions? \\

Reflection (RF) & Utterance that mirrors the client's thoughts, feelings, or experiences to show understanding. & It sounds like you're feeling overwhelmed with your current schedule. \\

Therapist Input (TI) & Any therapist's utterances are not coded as ``question'' or ``reflection''.
& Okay. Keep going. \\

\end{tabularx}
\begin{tabularx}{\textwidth}{>{\raggedright\arraybackslash}p{0.18\textwidth} >{\raggedright\arraybackslash}p{0.46\textwidth} >{\raggedright\arraybackslash}X}
\midrule \midrule
\textbf{\begin{tabular}[c]{@{}l@{}}Client Code\end{tabular}} & \textbf{Description} & \textbf{Example} \\

\midrule

Change Talk (CT) & Utterance indicate a desire, ability, reasons, or need for change.  & I really want to quit smoking because I want to be healthier. \\

Sustain Talk (ST) & Utterance argue against change or express a desire to maintain the status.  & I've tried quitting smoking before, but it's just too hard for me. \\

Neutral Talk (NT) & Utterance are neither for nor against change. & I've been smoking since I was a teenager. It’s been a part of my life for a long time. \\

\bottomrule
\end{tabularx}
\caption{The complete coarse-grained MISC codes in the AnnoMI dataset.}
\end{table*}


\begin{table*}[ht!]
\scriptsize
\renewcommand{\arraystretch}{1.60}
\begin{tabularx}{\textwidth}
{>{\raggedright\arraybackslash}p{0.18\textwidth} >{\raggedright\arraybackslash}p{0.46\textwidth} >{\raggedright\arraybackslash}X}
\toprule
\textbf{\begin{tabular}[c]{@{}l@{}}Therapist Code\end{tabular}} & \textbf{Description (abbreviated version)} & \textbf{Example} \\
\midrule
\Acf{OQ} & Asking questions for a wide range of answers. & Can you tell me more about your drinking habits? \\
\Acf{CQ} & Asking questions for concise answers: ``Yes'' or ``no'', a number. & Did you use heroin this week? \\
\midrule
\Acf{SR} & Conveying shallow understanding without additional information. & You don’t want to do that. \\
\Acf{CR} & Conveying deep understanding with additional information.
& That’s where you drew the line. \\
\midrule
\Acf{ADV} & Providing suggestions or recommendations. & Consider starting with small, manageable changes like taking a short walk daily. \\
\Acf{AFF} & Conveying positive or complimentary information. & You did well by seeking help. \\
\Acf{DIR} & Offering an imperative order, command, or direction. & You’ve got to stop drinking. \\
\Acf{EC} & Emphasizing client's freedom of choice. & It's up to you to decide whether to drink. \\
\Acf{FA} & Encouraging the client to keep sharing. & Tell me more about that. \\
\Acf{FIL} &  \done{Fitlering utterances are not related to behavior change.} & Good Morning! \\
\Acf{GI} & Offering relevant information, explanations, or feedback. & There are several treatment options available for managing stress. \\
\Acf{SP} & Offering encouragement and reassurance & I'm here to support you through your recovery journey.\\
\Acf{STR} & Offering a treatment process during the client's journey.  & First, let's discuss your drinking, and then we can explore other issues. \\
\Acf{WAR} & Offering a warning or negative consequences. & You could go blind if you don’t manage your blood sugar levels. \\
\Acf{PS} & Asking for consent before providing information or advice. & May I suggest a few stress management techniques? \\
\Acf{OP} & Expressing a viewpoint or judgment & In my opinion, addressing your stress can help reduce your drinking. \\
\end{tabularx}
\begin{tabularx}{\textwidth}{>{\raggedright\arraybackslash}p{0.18\textwidth} >{\raggedright\arraybackslash}p{0.46\textwidth} >{\raggedright\arraybackslash}X}
\midrule \midrule
\textbf{\begin{tabular}[c]{@{}l@{}}Client Code\end{tabular}} & \textbf{Description} & \textbf{Example} \\
\midrule
\Acf{FN} & No indication of client inclination toward or away from change. & Yeah. \\
\Acf{ASK} & Asking for clarification or information. & What treatment options are available? \\
\midrule

\Acl{CM} (\acs{CM+}/\acs{CM-}) & An agreement, intention, or obligation regarding future change. & I will try to reduce my drinking. \\

\Acl{TS} (\acs{TS+}/\acs{TS-}) & Concrete steps the client has recently taken to make a change. & I threw away all of my cigarettes. \\

\Acl{R} (\acs{R+}/\acs{R-}) & Rationale, basis, justification, or motive to make a change. & It would be so good for my kids. \\

\Acl{O} (\acs{O+}/\acs{O-}) & Other statements clearly reflect intention of change. & My family doesn’t believe I can quit. \\
\bottomrule
\end{tabularx}
\caption{{The complete fine-grained MISC codes in the \acl{BiMISC} dataset. The symbols ``+'' and ``-'' represent the client's desire to change (+) or not change (-) their behaviors with \acs{CM}, \acs{TS}, \acs{R} or \acs{O} intention.}}
\end{table*}
\clearpage

\begin{appendices}
\renewcommand{\thesection}{Appendix E}
\section{\done{Information letter and consent form}}
\label{consent}
\end{appendices}

\begin{table}[ht!]
\centering
\footnotesize
\renewcommand{\arraystretch}{1.2}
\begin{tabularx}{\textwidth}{>{\raggedright\arraybackslash}p{0.25\textwidth} >
{\raggedright\arraybackslash}X} 

\hline
\multicolumn{1}{l}{\textbf{\textcolor{red}{Expert Evaluation}}} & \multicolumn{1}{l}{\textbf{}} \\
\hline

\textbf{Information Letter} & 
Dear participant, \newline \newline 
Thank you for contributing to our study. This letter aims to provide you with essential information about the study's background and objectives. \newline \newline 
Background of the study \newline
Our research focuses on evaluating Artificial Intelligence (AI) generated dialogues within the context of Motivational Interviewing (MI) counseling.  \newline \newline 
What is this survey about?  \newline 
You are invited to participate in an online survey that will take approximately 45-60 minutes. 
The survey involves evaluating AI-generated dialogue responses in health counseling settings. Your insights are invaluable to understanding the effectiveness of these AI models. There are no right or wrong answers, so please respond based on your personal perspective. If you encounter any issues or have questions, please contact us at your convenience. \newline \newline 
Participation \newline 
Your participation is entirely voluntary. During the experiment, you are free to stop participating at any moment without giving a reason for doing so. No personal identifying information will be collected, and all demographic data will remain confidential. Only the research team will have access to the data, and any published results will be anonymized. Still, do not hesitate if you have any concerns, and please communicate with the responsible researcher if anything happens.  \newline \newline 
Further information \newline 
Should you have questions about this study at any given moment, please contact the responsible researcher(s): [anonymous]. Formal complaints about this study can be addressed to the Ethics Review Board; [anonymous]. For questions and formal complaints about the protection of your personal information, please contact the Data Protection Officer: [anonymous] \newline \newline 
Thank you, \newline 
\\ \hline

\textbf{Consent Form} &  
In this form, we refer to the information letter describing the research in which you participate. By signing this form, you declare that you understand the nature and methods of this study as described in the information letter. \newline \newline 
Should you have questions about this study at any given moment, please contact the responsible researchers: [anonymous]. Formal complaints about this study can be addressed to the Ethics Review Board: [anonymous]. For questions and formal complaints about the protection of your personal information, please contact the Data Protection Officer: [anonymous] \newline \newline 
By selecting ``Agre'' you confirm that:\newline 
• I am 18 years or older.\newline 
• I have read and understood the information letter.\newline 
• I agree to participate in the study and use the data obtained with it.\newline 
• I understand that I can withdraw the participation from the study at any moment without providing any reason. \newline 
\\ \hline 
\end{tabularx}
\caption{The information letter and consent form used for ``Expert Evaluation''.}
\end{table}


\begin{table}[ht!]
\centering
\footnotesize
\renewcommand{\arraystretch}{1.2}
\begin{tabularx}{\textwidth}{>{\raggedright\arraybackslash}p{0.25\textwidth} >
{\raggedright\arraybackslash}X} 
\hline
\multicolumn{1}{l}{\textbf{\textcolor{red}{Laypeople Evaluation}}} & \multicolumn{1}{l}{\textbf{}} \\
\hline
\textbf{Information Letter} & 
Dear participant, \newline \newline
Thank you for showing interest in our study. This letter aims to provide you with essential information about the study's background and objectives. \newline \newline
Background of the study \newline
Our research focuses on evaluating Artificial Intelligence (AI) generated dialogues within the context of Motivational Interviewing (MI) counseling. We aim to assess the performance of various AI models developed based on real-life therapeutic dialogues, with a particular emphasis on their application in health counseling scenarios, such as encouraging healthy behaviors. \newline \newline
What is this survey about? \newline
You are invited to participate in an online survey that will take approximately 30 minutes. The survey involves evaluating AI-generated dialogue responses in health counseling settings. Your insights are invaluable to understanding the effectiveness of these AI models. We recommend completing the survey in a quiet setting and focusing on the instructions and questions provided. There are no right or wrong answers, so please respond based on your personal perspective. If you encounter any issues or have questions, please contact us at your convenience. \newline \newline 
Participation \newline
Your participation is entirely voluntary. To ensure the quality of our data, responses completed in under 15 minutes will be considered invalid. As a token of our appreciation, you will receive a compensation of 4 Euros upon completion. \newline \newline
Discomfort, risks, and insurance \newline
We will try to minimize all potential risks during the lab procedures. During the experiment, you are free to stop participating at any moment without giving a reason for doing so. Still, do not hesitate if you have any concerns, and please communicate with the responsible researcher if anything happens. \newline \newline
Your privacy is guaranteed \newline
We assure you that your privacy is of utmost importance. No personal identifying information will be collected, and all demographic data will remain confidential. Only the research team will have access to the data, and any published results will be anonymized. \newline \newline
Further information \newline
Should you have questions about this study at any given moment, please contact the responsible researcher(s): [anonymous] \newline \newline
Thank you. \newline 
\\ \hline

\textbf{Consent Form} &  
In this form, we refer to the information letter describing the research in which you participate. By signing this form, you declare that you understand the nature and methods of this study as described in the information letter. \newline \newline 
Should you have questions about this study at any given moment, please contact the responsible researchers: [anonymous]. Formal complaints about this study can be addressed to the Ethics Review Board: [anonymous]. For questions and formal complaints about the protection of your personal information, please contact the Data Protection Officer: [anonymous] \newline \newline 
By selecting ``Agre'' you confirm that:\newline 
• I am 18 years or older.\newline 
• I have read and understood the information letter.\newline 
• I agree to participate in the study and use the data obtained with it.\newline 
• I understand that I can withdraw the participation from the study at any moment without providing any reason. \newline 
\\ \hline 
\end{tabularx}
\caption{The information letter and consent form used for ``Laypeople Evaluation''.}
\end{table}

\end{document}